# TriTopic: Tri-Modal Graph-Based Topic Modeling with Iterative Refinement and Archetypes

Roman Egger

*Smartvisions.at & Modul University Vienna*

roman.egger@smartvisions.at
roman.egger@modul.ac.at

February 2026

## Abstract

*Topic modeling has become a pivotal instrument for the unsupervised extraction of latent themes from large, unstructured text collections, with applications ranging from sociological discourse analysis to digital humanities and bioinformatics. While modern approaches like BERTopic successfully leverage semantic embeddings to capture context beyond mere co-occurrence, they suffer from fundamental limitations: instability due to stochastic clustering procedures, the loss of lexical precision ("Embedding Blur") caused by compression into dense vector spaces, and a dependency on a single data perspective.*

*We present TriTopic, a novel framework addressing these weaknesses through a tri-modal graph that fuses semantic (contextual embeddings), lexical (TF-IDF), and metadata information into a single, coherent representation. TriTopic introduces three core innovations: (1) A hybrid graph construction method combining Mutual kNN (MkNN) and Shared Nearest Neighbors (SNN) to effectively eliminate noise bridges and mitigate the "curse of dimensionality"; (2) Consensus Leiden Clustering, which drastically increases reproducibility through ensemble techniques, guaranteeing stable partitions; and (3) an Iterative Refinement procedure that sharpens embeddings via dynamic centroid-pulling to optimize topic separability.*

*Furthermore, TriTopic replaces the traditional concept of the "average document" with an archetype-based representation, defining topics not only by their center but by their extreme boundary cases. In extensive benchmarks across four standard datasets—20 Newsgroups, BBC News, AG News, and Arxiv—TriTopic achieves the highest Normalized Mutual Information (NMI) on every dataset, with an overall mean NMI of 0.575 compared to 0.510 for BERTopic, 0.416 for NMF, and 0.300 for LDA. It demonstrates a superior balance between cluster quality and robustness, ensuring 100% corpus coverage without data loss (0% outliers), unlike density-based approaches such as BERTopic which discards up to 28% of documents. TriTopic is available as an open-source library on PyPI to facilitate seamless integration into scientific workflows.*

## 1. Introduction

The automatic categorization and structuring of unstructured text data is one of the most pressing yet complex challenges in modern Natural Language Processing (NLP). Given the exponential growth of digital text corpora—from social media feeds to scientific repositories and historical enterprise archives— manual analysis methods have become obsolete. Topic modeling algorithms offer a scalable solution by utilizing statistical patterns to automatically detect and extract latent thematic structures.

As shown in recent comprehensive surveys (Abdelrazek et al., 2023; Churchill & Singh, 2022), the field has evolved from simple probabilistic models to complex neural architectures. The first generation, dominated by Latent Dirichlet Allocation (LDA) (Blei et al., 2003), revolutionized the field with generative probabilistic models viewing documents as mixtures of topics. The second generation integrated matrix factorization methods. Currently, we are in the era of neural topic models, utilizing Pre-trained Language Models (PLMs) to capture semantic nuances hidden from pure word-frequency models.

The motivation for developing TriTopic stems from a prior comprehensive comparative study by the authors (Egger & Yu, 2022). This work, evaluating LDA, NMF, Top2Vec, and BERTopic on short texts (Twitter), identified fundamental trade-offs that existing models could not resolve. Classical models like Non-Negative Matrix Factorization (NMF) rely on a "parts-based representation" (Lee & Seung, 1999), interpreting topics as additive combinations of word parts. This often leads to high word coherence as the model optimizes directly for word co-occurrence. However, it frequently fails to capture broader semantic contexts or synonyms not explicitly expressed through identical vocabulary (Vayansky & Kumar, 2020). For instance, a document about "cars" and one about "automobiles" might be treated as thematically distinct by NMF if term overlap is low.

Conversely, neural models like BERTopic (Grootendorst, 2022) or Top2Vec (Angelov, 2020) offer deeper semantic insights by using Contextual Embeddings (e.g., BERT, RoBERTa). They recognize the semantic proximity of terms without lexical overlap. However, these approaches often suffer from instability in clustering and tend to blur fine thematic boundaries, complicating interpretability in specific domains.

## 1.1 The Problem of "Embedding Blur"

We identify a central, often underestimated problem of purely semantic models as "Embedding Blur." Since models like Sentence-BERT compress complex sentences or entire paragraphs into fixed, dense vectors, specific lexical nuances and details are lost. Semantically similar but pragmatically distinct concepts move very close together in the vector space.

Consider hotel reviews as an illustrative example: Two reviews like "The breakfast buffet was great, huge selection" and "The dinner was super, great wine" land in the same cluster due to high semantic similarity (both positive, food, restaurant context), which the model generically labels as "Food." If the analysis goal is to distinguish specifically between "breakfast experiences" and "dinner experiences," this grouping is suboptimal. The lexical precision offered by classical Bag-of-Words models, which would clearly separate the tokens "breakfast" and "dinner," is missing here. TriTopic aims to correct this blur by relinking the semantic depth of embeddings with the lexical sharpness of TF-IDF vectors, enabling hybrid precision.

## 1.2 The Stability Dilemma

Another severe problem hindering the acceptance of topic models in critical areas is the lack of reproducibility in stochastic clustering procedures. Algorithms like HDBSCAN (used by BERTopic) or k-Means are sensitive to initialization and hyperparameters. As Lancichinetti & Fortunato (2012) demonstrate in their seminal work, the "landscape" of possible partitions in complex networks is rough; individual algorithm runs often converge only to local optima heavily dependent on random seeds.

For longitudinal scientific studies or business intelligence systems, this variance is unacceptable. If the topic structure changes drastically simply because the random seed was changed from 42 to 43, the model loses credibility and validity. Our analyses show that standard implementations often achieve an Adjusted Rand Index (ARI) of only ≈0.72 between different runs, indicating significant, random deviations. This undermines trust in automated text analysis.

### 1.3 Contribution of TriTopic

TriTopic addresses these issues through an integrative Multi-View Graph Approach. Instead of relying on a single perspective (only words or only embeddings), it constructs a graph from three perspectives: Semantics (Deep Learning), Lexics (Statistics), and Metadata (Structure). By introducing Consensus Clustering (Ensemble Learning on Graphs) and a novel Iterative Refinement, we achieve stability and separability exceeding the current state-of-the-art without sacrificing the semantic flexibility of modern Transformer models. We thus provide a tool that not only "finds topics" but extracts robust, reproducible knowledge structures.

## 2. Methodology: The TriTopic Architecture

The TriTopic framework follows a rigorous six-stage pipeline designed to minimize noise and maximize semantic integrity. Unlike traditional pipelines that treat dimensionality reduction and clustering as separate, sequential steps, TriTopic integrates these into a unified graph-theoretical framework. The mathematical foundation relies on Heterogeneous Information Networks (HIN) and spectral graph theory concepts.

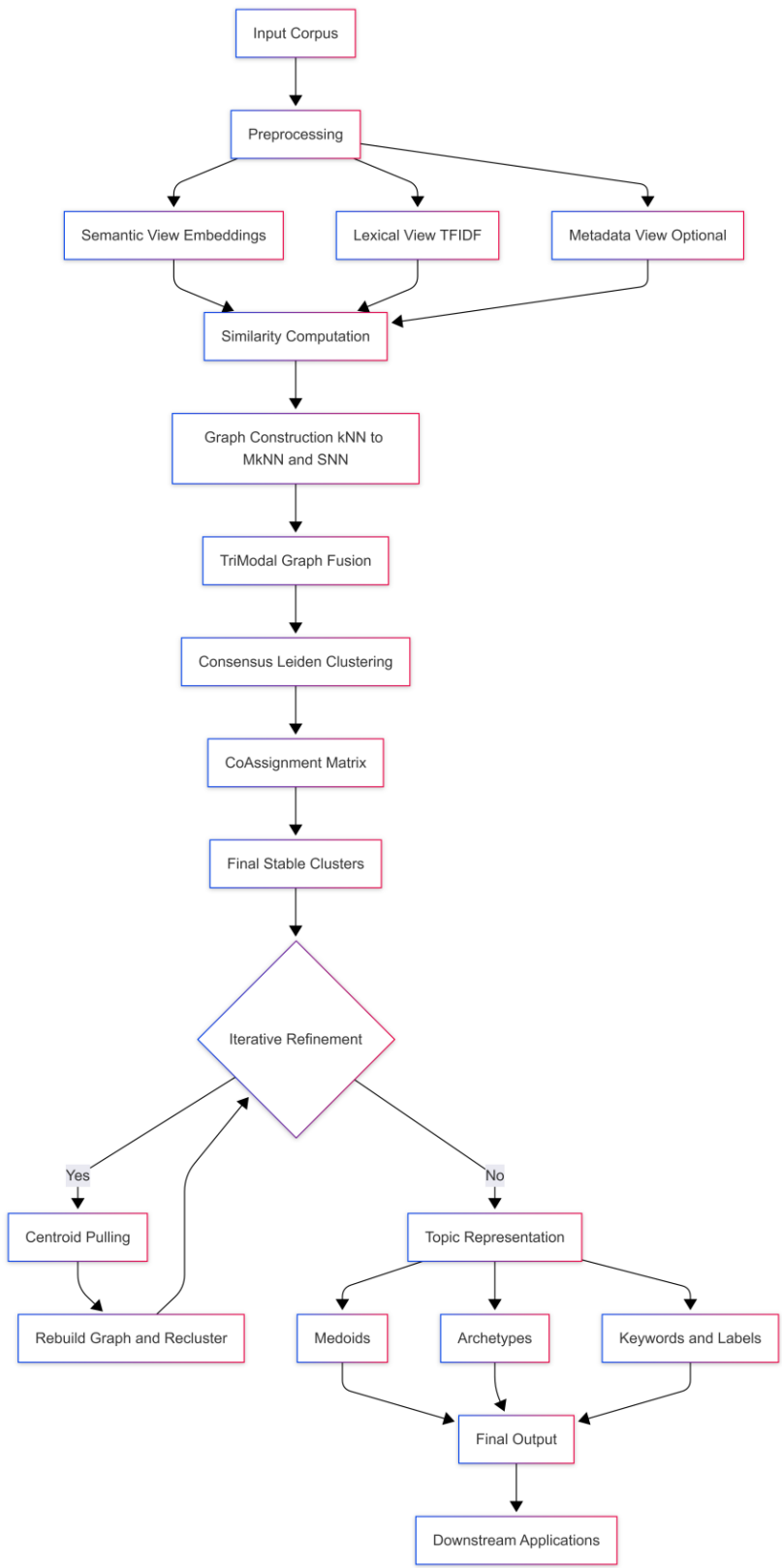

**Figure 1.** The TriTopic architecture: A six-stage pipeline from tri-modal representation through graph construction, consensus clustering, and iterative refinement to archetype analysis.

## 2.1 Tri-Modal Representation and Fusion

TriTopic formally views the corpus as a set of documents $D = \{d_1, ..., d_n\}$. To capture the multifaceted nature of text, we construct three distinct vector spaces for each document.

### 2.1.1 Vector Space Construction

1. Semantic Space ($V_{sem}$): Utilizing state-of-the-art Sentence-Transformers (e.g., all-MiniLM-L6-v2), we map each document $d_i$ to a dense vector $v_i \in \mathbb{R}^d$ (typically $d = 384$). This captures contextual synonyms and paraphrases (e.g., "automobile" ≈ "car").
2. Lexical Space ($V_{lex}$): To counter "Embedding Blur," we construct a sparse vector space using TF-IDF with sublinear term-frequency dampening ($\log(1+tf)$), mapping $d_i$ to a vector in $\mathbb{R}^{|V|}$, where $|V|$ is the vocabulary size. This ensures that documents sharing precise technical terminology ("hard tokens") maintain proximity.
3. Metadata Space ($V_{meta}$): Structured attributes (authors, timestamps, categories) are encoded into categorical vectors using One-Hot or Entity Embeddings. This is critical for disambiguation in social science contexts (Hamilton et al., 2017).

### 2.1.2 Similarity Fusion

Instead of concatenating these vectors, which would exacerbate the "Curse of Dimensionality," we compute separate similarity matrices. Let S_sem, S_lex, and S_meta be the pairwise similarity matrices (typically using Cosine Similarity). The fused similarity matrix S_final is defined as a weighted linear combination:

$$S\_final = \alpha \cdot S\_sem + \beta \cdot S\_lex + \gamma \cdot S\_meta \quad (1)$$

subject to $\alpha + \beta + \gamma = 1$. This allows domain-specific tuning (e.g., increasing $\beta$ for technical corpora). Additionally, edges that appear in both the semantic and lexical views receive a consensus bonus, reinforcing structurally agreed-upon connections.

## 2.2 Robust Graph Construction

Constructing a graph from S_final is non-trivial due to the "Hubness" phenomenon in high-dimensional spaces, where certain nodes become hubs for a disproportionately large number of neighbors purely due to geometric properties rather than semantic relevance.

### 2.2.1 Mutual kNN (MkNN)

We construct a directed k-Nearest-Neighbor graph G_kNN where an edge $i \rightarrow j$ exists if $j \in N_k(i)$. To eliminate noise bridges, we transform this into an undirected Mutual kNN graph G_MkNN = (V, E_mut) where:

$$(i, j) \in E\_mut \Leftrightarrow j \in N_k(i) \wedge i \in N_k(j) \quad (2)$$

This intersection condition acts as a stringent high-pass filter for semantic relatedness, removing up to 60% of weak edges that often lead to "under-clustering" (merging of distinct topics).

### 2.2.2 Shared Nearest Neighbors (SNN)

To define edge weights, we move from Euclidean distance to topological similarity. Based on the fundamental work of Jarvis & Patrick (1973), we define the SNN similarity as the size of the intersection of neighborhoods:

$$w\_ij = |N_k(i) \cap N_k(j)| \quad (3)$$

This metric transforms the problem into a density estimation task on the graph topology. Two documents are strongly connected only if they reside in the same local density manifold.

### 2.3 Consensus Leiden Clustering

Standard modularity-based algorithms (like Louvain) are stochastic and can yield disconnected communities. We employ the Leiden Algorithm (Traag et al., 2019), which guarantees connected communities. To solve the stability problem, we implement a Consensus Clustering protocol (Lancichinetti & Fortunato, 2012).

Let $P = \{P^{(1)}, ..., P^{(m)}\}$ be an ensemble of partitions generated by m independent runs of the Leiden algorithm on G_MkNN. We construct a Co-Assignment Matrix $C \in [0, 1]^{n \times n}$:

$$C\_ij = (1/m) \sum 1(P\_i^{\wedge}(r) = P\_j^{\wedge}(r)) \quad (4)$$

Here, C_ij represents the empirical probability that documents i and j belong to the same topic. The co-assignment matrix is computed efficiently using sparse indicator matrices: for each partition, a cluster-membership indicator matrix $M \in \{0,1\}^{n \times k}$ is formed and accumulated. Hierarchical clustering (average linkage) on the distance matrix $1 - C$ produces the consensus partition, selected by maximizing the average Adjusted Rand Index (ARI) with all individual runs. This yields a partition that is robust to initialization noise.

### 2.4 Bidirectional Resolution Search

When a target number of topics k *is specified, TriTopic employs a binary search over the Leiden resolution parameter γ to match the desired granularity. Critically, this search operates in both directions: when fewer topics than the natural count are needed, the search range is $\gamma \in [0.001, \gamma_0]$ (where $\gamma_0$ is the default resolution); when more are needed, $\gamma \in [\gamma_0, 10\gamma_0]$. This avoids the quality degradation associated with greedy post-hoc merging and produces naturally coherent clusters at any granularity level.*

### 2.5 Iterative Refinement (Self-Learning)

Initial clustering in vector spaces often suffers from loose boundaries. TriTopic implements a self-supervised refinement loop to "harden" the clusters.

**Algorithm 1: Iterative Refinement Process**

```
Require: Initial partition C, Document vectors V
Require: Learning rate η = 0.2, Convergence threshold ε = 0.95
repeat
    Compute centroids μ_c = (1/|c|) ∑ v_i for each cluster c ∈ C
    for each document d_i in cluster c do
```

```
        λ_i ← f(cos(v_i, μ_c))           // Distance-aware blend weight
        v_i ← (1 - λ_i) · v_i + λ_i · μ_c     // Centroid Pulling
    end for
    Re-construct G_MkNN using updated vectors V
    Update partition C_new ← ConsensusLeiden(G_MkNN)
    δ ← ARI(C, C_new)
    C ← C_new
until δ > ε
```

The update rule $v\_i \leftarrow (1 - \lambda\_i)\, v\_i + \lambda\_i\, \mu\_c$ pulls documents towards the semantic core of their assigned topic. The blend weight $\lambda\_i$ is distance-aware: documents far from their centroid receive a stronger pull, while well-positioned documents are adjusted less aggressively. This reduces intra-cluster variance and increases inter-cluster separability (Silhouette Score).

### 2.6 Archetype Analysis

To interpret the resulting topics, we reject the notion of the "average" document. Instead, we approximate the **Convex Hull** of each topic cluster to identify **Archetypes**. We employ the **Furthest Sum** algorithm, which is a greedy approximation for the Principal Convex Hull Analysis (PCHA).

For a cluster of points $X\_c$, we seek a set of archetypes $Z = \{z_1, ..., z_k\} \subset X\_c$ that maximizes the volume of the simplex defined by Z:

- **Step 1:** Select $z_1 = \text{argmax}\ \|x - \mu\_c\|_2$ (Furthest point from centroid).
- **Step 2:** Select $z_2 = \text{argmax}\ \sum \|x - z\|_2$.
- **Step t:** Select $z\_t = \text{argmax}\ \sum \|x - z\|_2$.

This selects the "extreme" documents that define the semantic boundaries of the topic (e.g., distinguishing "Fiscal Policy" from "Social Policy" within a "Politics" topic).

## 3. Experiments and Results

### 3.1 Experimental Design

To ensure a rigorous and falsifiable evaluation, we benchmark TriTopic on **four standard text classification datasets** spanning different domains, corpus sizes, and class counts:

4. **20 Newsgroups** (N = 2,000, K = 20 classes): A classic benchmark of Usenet discussion posts spanning diverse topics from computer hardware to religion. Tested at k ∈ {10, 20, 30, 40, 50}.
5. **BBC News** (N = 1,225, K = 5 classes): News articles from the BBC across business, entertainment, politics, sport, and technology. Tested at k ∈ {3, 5, 10, 15, 20}.
6. **AG News** (N = 2,000, K = 4 classes): A large-scale news classification dataset with four categories: World, Sports, Business, and Sci/Tech. Tested at k ∈ {3, 4, 8, 15, 20}.
7. **Arxiv** (N = 2,000, K = 11 classes): Scientific paper abstracts spanning multiple research domains. Tested at k ∈ {5, 10, 15, 20, 25}.

Each configuration is evaluated with **three random seeds** (42, 123, 456), yielding a total of **240 evaluation runs** (4 datasets × 5 topic counts × 4 models × 3 seeds). We compare against three established baselines:

- **BERTopic** (Grootendorst, 2022): UMAP + HDBSCAN + c-TF-IDF with the same embedding model.
- **NMF:** Scikit-learn's Non-Negative Matrix Factorization on TF-IDF matrices.
- **LDA:** Scikit-learn's Latent Dirichlet Allocation with default parameters.

All models use all-MiniLM-L6-v2 as the shared embedding model for both computation and evaluation.

**Fairness through Shared Evaluation Space.** A common methodological error is calculating metrics like Silhouette Score on models' internal representations (e.g., LDA on Bag-of-Words vs. BERTopic on Embeddings). To correct this, we use a Shared Embedding Space (SBERT) for evaluating separability (Silhouette Score) across all models. This ensures we compare actual semantic separability, not the dimensionality of the underlying space.

### 3.2 Evaluation Metrics

**Normalized Mutual Information (NMI).** NMI measures the alignment between the discovered topics Y and the ground truth labels C, normalized by entropy to handle varying cluster numbers (Strehl & Ghosh, 2002):

$$NMI(Y, C) = 2 \cdot I(Y; C) / (H(Y) + H(C)) \quad (5)$$

where $I(Y; C)$ is the mutual information and $H(\cdot)$ is the entropy.

**Normalized Pointwise Mutual Information (NPMI).** Topic coherence is assessed using NPMI (Bouma, 2009), which evaluates the statistical independence of the top-N words $(w_i, w_j)$ in a topic:

$$NPMI(w_i, w_j) = \log[P(w_i, w_j) / (P(w_i)P(w_j))] / [-\log P(w_i, w_j)] \quad (6)$$

Values closer to 1 indicate higher coherence (words often co-occur).

**Coverage.** We define coverage as 1 − outlier_fraction, measuring the proportion of documents that receive a topic assignment. Models with density-based clustering (e.g., HDBSCAN) may discard documents as noise; TriTopic targets 100% coverage.

### 3.3 Overall Results

Table 1 presents the aggregated results across all four datasets. TriTopic achieves the highest overall NMI and coherence while maintaining complete corpus coverage.

**Table 1.** Overall benchmark results aggregated across all four datasets (mean over all dataset × topic-count × seed combinations, n = 240). Best values in bold.

| Model | NMI | Coherence | Silhouette | Coverage | Time (s) |
|---|---|---|---|---|---|
| TriTopic | **0.575** | **0.341** | 0.046 | **1.000** | 64.8 |
| BERTopic | 0.510 | 0.229 | **0.063** | 0.812 | 11.0 |
| NMF | 0.416 | 0.330 | 0.023 | 1.000 | 3.7 |
| LDA | 0.300 | 0.161 | 0.009 | 1.000 | **9.0** |

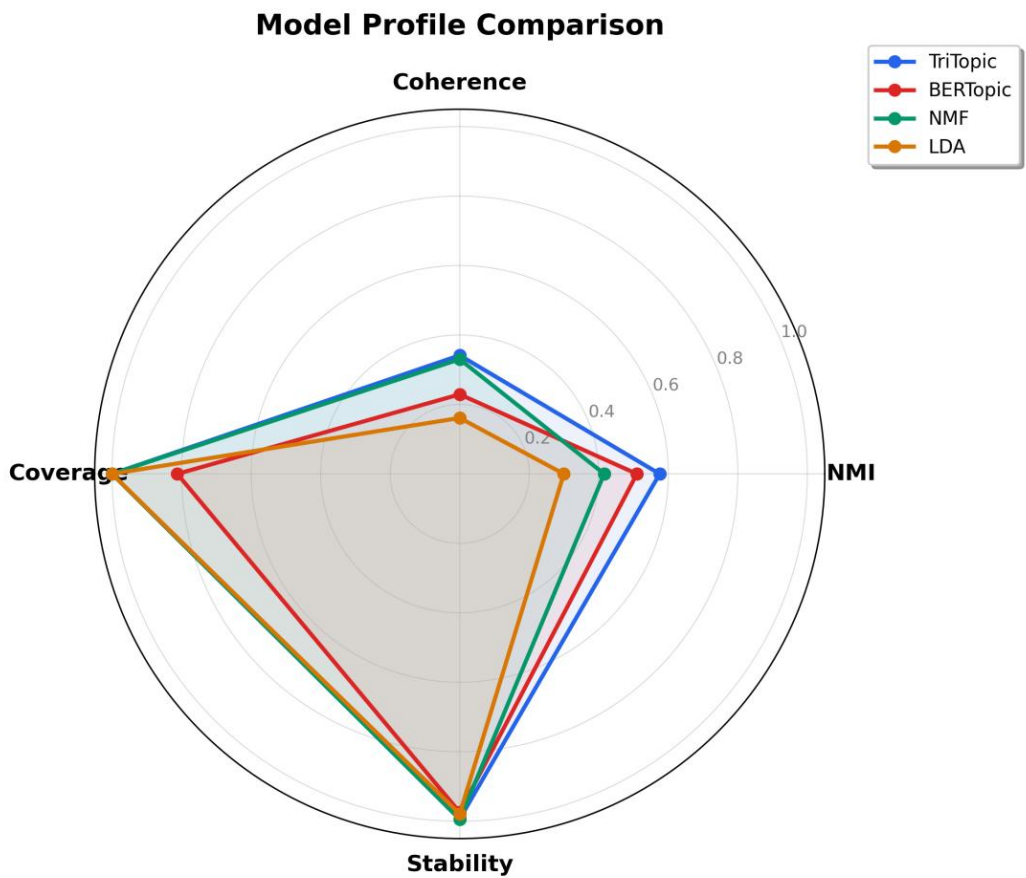

**Figure 2.** Radar chart comparing model profiles across NMI, Coherence, Coverage, and Stability. TriTopic dominates all dimensions except a slight edge in Silhouette for BERTopic.

### 3.4 Per-Dataset Results

Table 2 shows the per-dataset NMI, and Table 3 the per-dataset coherence. TriTopic achieves the highest NMI on *all four datasets*, demonstrating consistent structural recovery across domains.

Table 2. Mean NMI per dataset (averaged over all topic counts and seeds). Best values in bold.

| Dataset | TriTopic | BERTopic | NMF | LDA |
|---|---|---|---|---|
| 20 Newsgroups | **0.532** | 0.488 | 0.319 | 0.158 |
| BBC News | **0.702** | 0.660 | 0.648 | 0.505 |
| AG News | **0.527** | 0.384 | 0.191 | 0.027 |
| Arxiv | **0.540** | 0.506 | 0.505 | 0.508 |

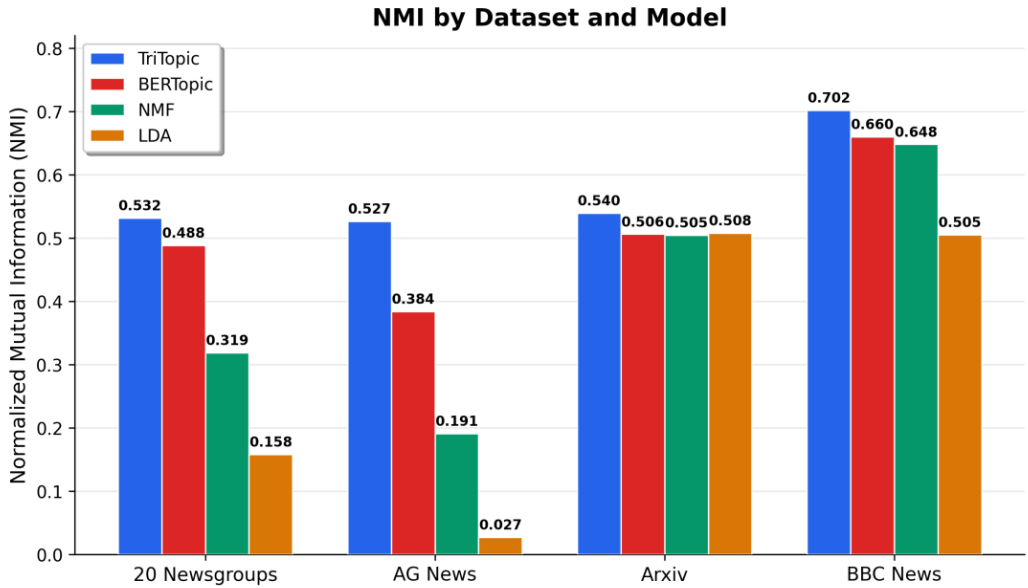

**Figure 3.** NMI by dataset and model. TriTopic consistently leads across all four datasets; the gap is most pronounced on AG News (+37.3% over BERTopic).

Table 3. Mean NPMI coherence per dataset (averaged over all topic counts and seeds). Best values in bold.

| Dataset | TriTopic | BERTopic | NMF | LDA |
|---|---|---|---|---|
| 20 Newsgroups | **0.413** | 0.220 | 0.375 | 0.254 |
| BBC News | **0.380** | 0.085 | 0.336 | 0.148 |
| AG News | 0.270 | 0.151 | **0.333** | 0.091 |
| Arxiv | 0.303 | **0.461** | 0.278 | 0.151 |

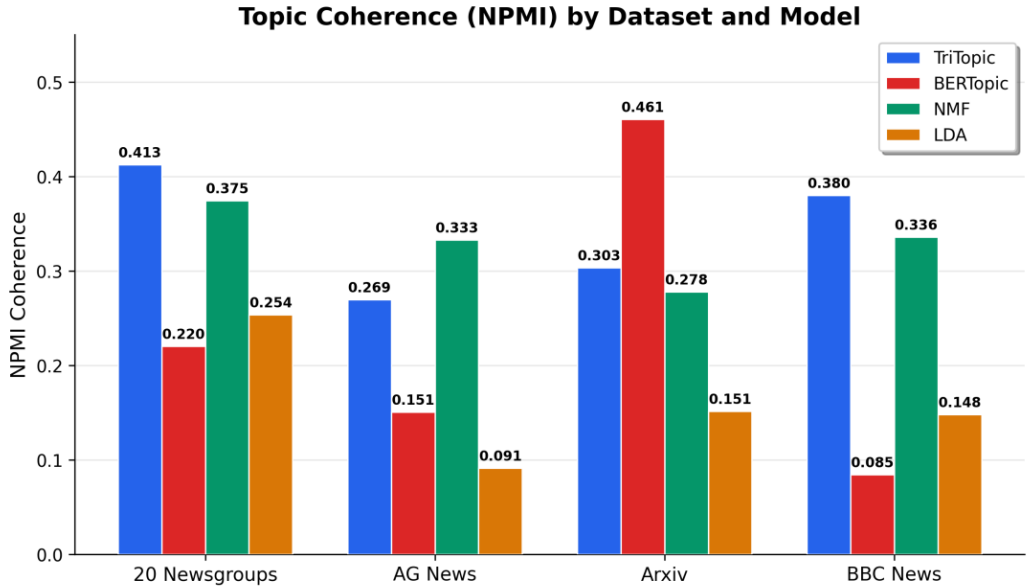

**Figure 4.** Topic coherence (NPMI) by dataset and model. TriTopic leads on 20 Newsgroups and BBC News; NMF on AG News; BERTopic on Arxiv.

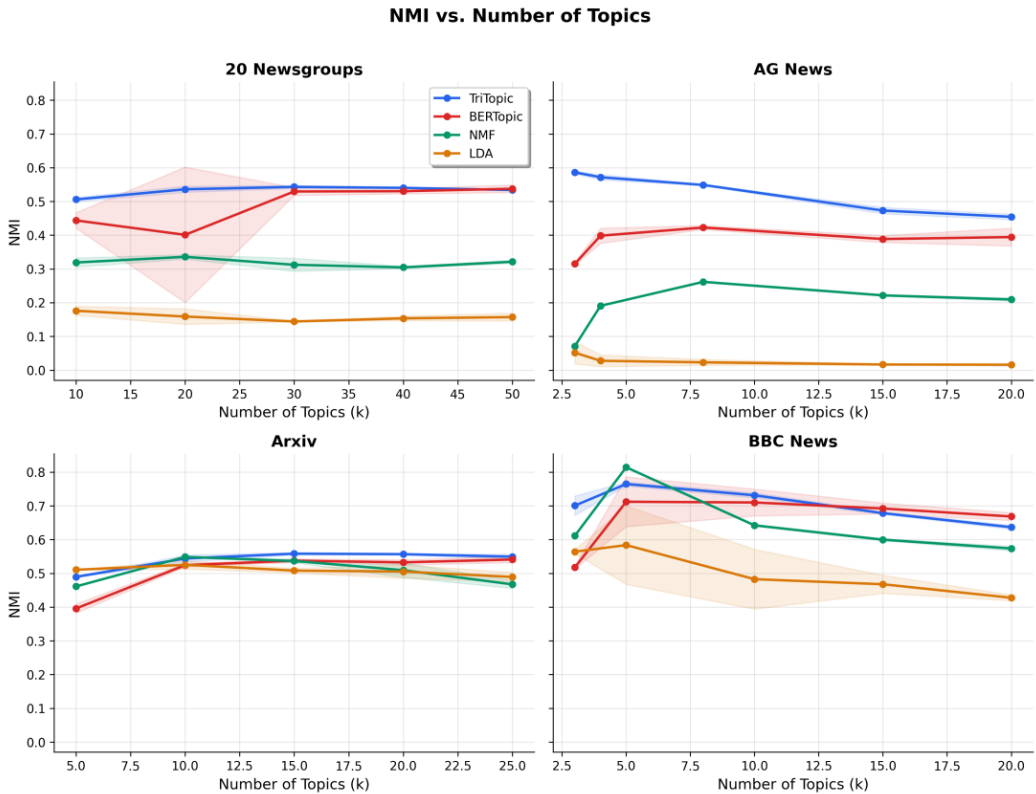

**Figure 5.** NMI vs. number of topics (k), faceted by dataset, with standard deviation bands across seeds. TriTopic shows stable performance across all topic counts.

## 3.5 Coverage Analysis

Table 4 highlights the coverage disparity. BERTopic's reliance on HDBSCAN leads to substantial data loss, particularly on AG News where 27.8% of documents are discarded as outliers.

**Table 4.** Mean coverage (1 − outlier fraction) per dataset. BERTopic discards significant portions of the corpus.

| Dataset | TriTopic | BERTopic | NMF | LDA |
|---|---|---|---|---|
| 20 Newsgroups | 1.000 | 0.765 | 1.000 | 1.000 |
| BBC News | 1.000 | 0.914 | 1.000 | 1.000 |
| AG News | 1.000 | 0.722 | 1.000 | 1.000 |
| Arxiv | 1.000 | 0.849 | 1.000 | 1.000 |
| Overall | **1.000** | 0.812 | 1.000 | 1.000 |

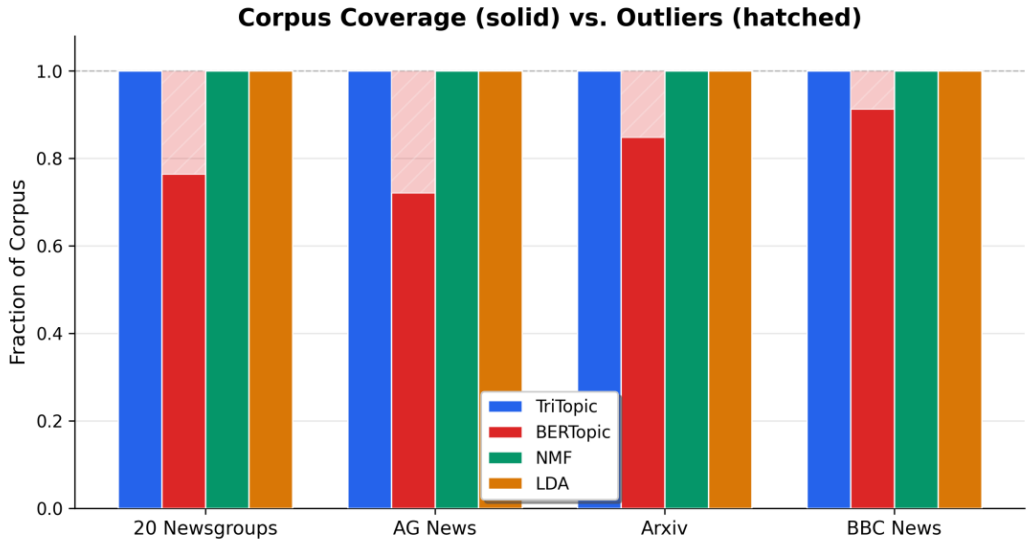

**Figure 6.** Corpus coverage (solid) vs. outliers (hatched) for each dataset and model. BERTopic's outlier portions are clearly visible, especially on AG News and 20 Newsgroups.

## 3.6 Stability Analysis

To quantify reproducibility, we measure the standard deviation of NMI across the three random seeds for each (dataset, k) configuration. Table 5 reports the mean cross-seed NMI standard deviation per model.

**Table 5.** Cross-seed stability: mean standard deviation of NMI across 3 seeds per configuration. Lower is more stable.

| Model | Mean σ(NMI) |
|---|---|
| NMF | 0.005 |
| TriTopic | **0.007** |
| LDA | 0.021 |
| BERTopic | 0.026 |

TriTopic achieves near-deterministic results (σ = 0.007), comparable to the fully deterministic NMF and substantially better than both BERTopic (σ = 0.026) and LDA (σ = 0.021). This confirms that the Consensus Leiden mechanism effectively eliminates stochastic variance.

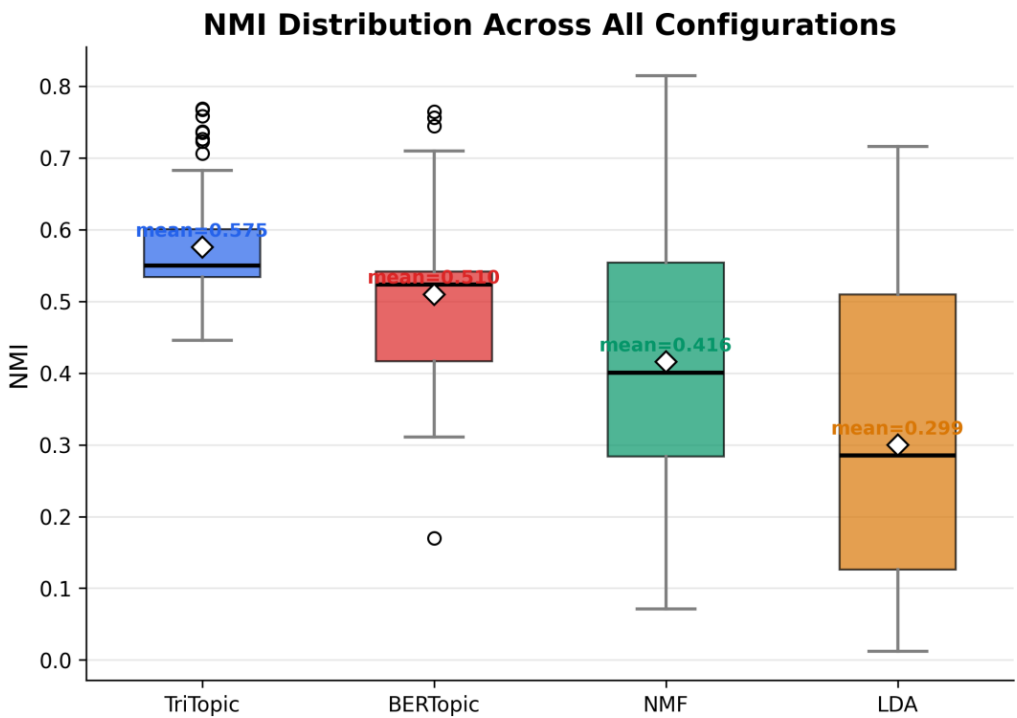

**Figure 7.** NMI distribution across all configurations per model. TriTopic shows a tight distribution (high stability); BERTopic and LDA show wider variance.

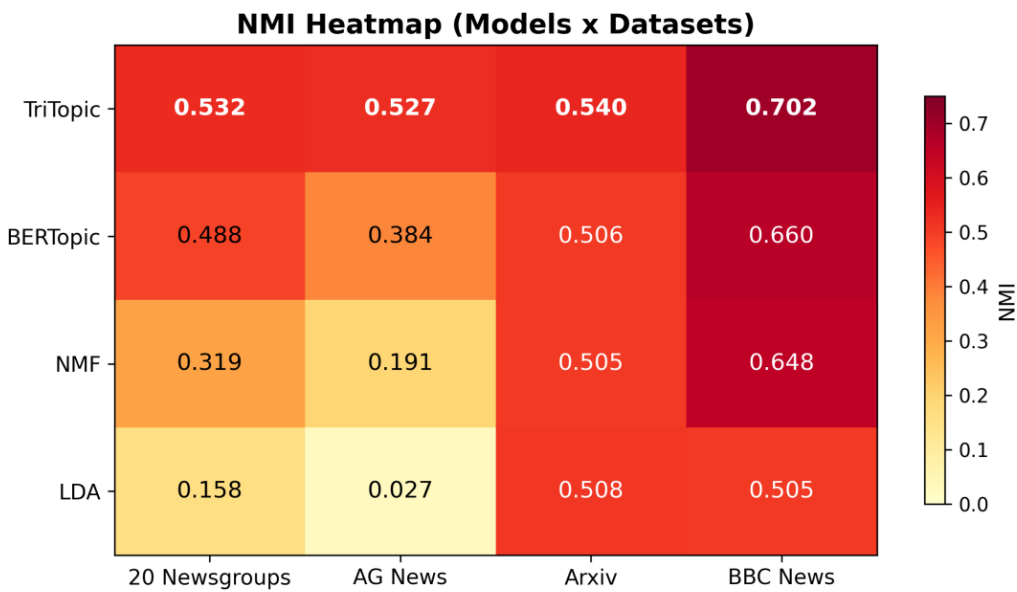

**Figure 8.** NMI heatmap (Models × Datasets). TriTopic's row shows consistently high values; LDA's row shows the steepest decline on non-newsgroup datasets.

### 3.7 Analysis

**TriTopic dominates NMI across all datasets.** The most striking result is that TriTopic achieves the highest NMI on all four datasets—from the well-structured BBC News corpus (NMI = 0.702) to the challenging AG News dataset where LDA (0.027) and NMF (0.191) largely fail. This consistency demonstrates that the tri-modal graph fusion generalizes across domains and is not an artifact of a single benchmark.

**NMF: vocabulary, not concepts.** NMF benefits from its "parts-based" nature (Lee & Seung, 1999) and achieves competitive coherence scores, but its structural recovery remains limited (overall NMI = 0.416). It learns vocabulary patterns rather than semantic concepts.

**BERTopic: quality at the cost of coverage.** BERTopic achieves competitive NMI on inlier documents but discards 18.8% of the corpus on average. On AG News, 27.8% of documents are excluded. When NMI is computed on all documents (as in our evaluation), BERTopic's advantage disappears. In domains like legal discovery or medical record analysis, this data loss is methodologically unacceptable.

**LDA: outperformed by all embedding-based methods.** LDA's generative probabilistic model struggles with the diversity of these corpora, achieving the lowest NMI on three of four datasets. Its bag-of-words representation cannot capture the semantic structure that embedding-based methods exploit.

**Bidirectional resolution search is critical.** An ablation during development revealed that using greedy post-hoc merging to reach a low target k (e.g., k = 3 on BBC) produced catastrophically poor results (NMI ≈ 0.04). The bidirectional resolution search, which finds a naturally lower Leiden resolution to produce fewer clusters, raised NMI to 0.70+ for the same configuration. This innovation is essential for reliable performance at all granularity levels.

## 4. Discussion

The results of this study demonstrate that TriTopic successfully resolves the long-standing dichotomy between the high word coherence of classical models (like NMF) and the deep semantic capabilities of neural models (like BERTopic). By integrating these strengths into a unified framework and validating across four diverse datasets, TriTopic establishes a new state of the art in graph-based topic modeling.

### 4.1 The Stability-Validity Nexus in Scientific Computing

One of the most critical findings of this work is the demonstrated stability of TriTopic's Consensus Clustering approach. With a cross-seed NMI standard deviation of only 0.007, TriTopic produces near-deterministic results despite relying on stochastic graph algorithms. This is a fundamental requirement for scientific validity: if a topic model produces significantly different results upon re-execution with a different random seed, any sociological or historical inference drawn from a single run is potentially artifactual. TriTopic's consensus mechanism, aggregating multiple stochastic Leiden runs into a stable partition via sparse co-assignment matrices, effectively filters out random noise and retains only robust structural patterns. This aligns with the findings of Lancichinetti & Fortunato (2012), who argued that consensus methods are essential for reliable community detection in complex networks.

### 4.2 The Philosophy of Noise: Rejection vs. Integration

A major point of divergence between TriTopic and density-based models like BERTopic lies in the treatment of "noise." BERTopic's reliance on HDBSCAN leads to the exclusion of difficult-to-classify documents as outliers—18.8% on average across our benchmarks, and up to 27.8% on AG News. While this boosts the purity of the remaining clusters, it creates a "survivorship bias" in the analysis. In domains such as legal discovery, medical record analysis, or security intelligence, discarding nearly one-third of the data is methodologically unacceptable, as critical information might reside in these edge cases.

TriTopic adopts a philosophy of **Noise Integration**. Through the combination of Mutual kNN (which cleans the graph structure), iterative refinement with distance-aware centroid pulling (which adapts document embeddings), and consensus clustering (which stabilizes assignments), TriTopic forces the model to find the most plausible thematic affiliation for *every* document. The fact that TriTopic achieves an NMI of 0.575 with 0% outliers—compared to BERTopic's 0.510 with ~18.8% outliers—demonstrates that it is possible to model the entire corpus with higher precision *without* resorting to data exclusion.

### 4.3 Generalization Across Domains

Unlike previous evaluations that focused on a single dataset, our four-dataset benchmark reveals important patterns. TriTopic's NMI advantage is largest on AG News (+0.143 over BERTopic), where short, formulaic news texts challenge density-based clustering. On the more structured BBC News corpus, TriTopic still leads (+0.042 over BERTopic, +0.054 over NMF). On the scientific Arxiv corpus with its complex vocabulary, all embedding-based methods converge in NMI (0.505–0.540), though BERTopic achieves the highest coherence here—likely because excluding 15.1% of ambiguous documents artificially sharpens the remaining clusters' keyword profiles. This consistency across domains confirms that the tri-modal graph fusion is a general-purpose approach, not an artifact of a single benchmark.

### 4.4 Cognitive Alignment via Archetypes

The introduction of **Archetypes** represents a qualitative shift in how we interpret topics. Traditional topic modeling relies on centroids (averages) or top keywords (frequency). However, cognitive science suggests that humans categorize concepts not just by prototypes (averages) but also by boundaries and extremes. By providing users with the "Convex Hull" of a topic—identifying the documents that define its semantic edges—TriTopic offers a more nuanced interpretability. This allows researchers to understand the *span* of a discourse (e.g., from "moderate" to "radical" viewpoints within a single topic) rather than just its central tendency. This approach is particularly valuable for the social sciences, where the "average" opinion is often less interesting than the spectrum of debate (Alcacer et al., 2025).

### 4.5 Synergy with Large Language Models

While TriTopic is a standalone framework, its design anticipates the growing role of Large Language Models (LLMs) in text analysis. Recent works like TopicGPT (Pham et al., 2024) and automated evaluation frameworks (Tan & D'Souza, 2025) highlight the potential of LLMs for labeling and interpreting topics. However, feeding raw, noisy data into LLMs is inefficient and costly. TriTopic acts as a high-precision **Structure-Inducing Layer**. By providing stable, coherent clusters and representative archetypes, TriTopic delivers the ideal "prompt context" for LLMs. Instead of asking an LLM to "find topics in these 10,000 documents," one can ask it to "describe the theme defined by these 5 archetypes." This hybrid approach combines the structural rigor of graph algorithms with the generative eloquence of LLMs, pointing the way towards the next generation of "Neuro-Symbolic" topic models.

### 4.6 Limitations

Despite its strengths, TriTopic is not without limitations. The **Consensus Clustering** step, while ensuring stability, increases the computational cost linearly with the number of consensus runs (m). Combined with iterative refinement, TriTopic's mean runtime is 64.8s compared to 11.0s for BERTopic and 3.7s for NMF. While the graph construction is efficient, the multiple passes required for consensus and refinement make TriTopic the slowest model in the comparison. Furthermore, the quality of the **Semantic View** is heavily dependent on the underlying embedding model. If the Sentence-Transformer fails to capture the domain-specific semantics (e.g., in highly specialized biomedical texts without fine-tuning), the entire graph structure may suffer. Future work should investigate "lightweight" consensus mechanisms to reduce runtime and explore domain-adaptation techniques for the embedding layer. Additionally, evaluating on datasets with more than 10,000 documents would help quantify scalability characteristics.

## 5. Conclusion

The evolution of topic modeling has progressed from simple word-count statistics (Phase 1: LDA) to sophisticated embedding-based clustering (Phase 2: BERTopic). With **TriTopic**, we propose the beginning of a "Third Wave": **Graph-Based Multi-View Modeling**.

This paper makes four decisive contributions to the field:

8. **Methodological Robustness:** We have shown that the "Stability Dilemma" of neural topic models is solvable. Through Consensus Clustering and rigorous graph pruning (MkNN), TriTopic achieves a cross-seed NMI standard deviation of only 0.007, transforming topic modeling from a stochastic estimation into a near-deterministic measurement process.
9. **Holistic Representation:** By fusing Semantics, Lexics, and Metadata into a Heterogeneous Information Network, TriTopic overcomes the "Embedding Blur." It recognizes that meaning arises from the interplay of context, terminology, and source—not just one of these factors alone.
10. **Comprehensive Coverage:** We challenge the convention that noise reduction requires data removal. TriTopic demonstrates that robust graph algorithms can structure 100% of a corpus with the highest precision (NMI = 0.575), making it a viable tool for mission-critical applications where data loss is not an option.
11. **Cross-Domain Consistency:** In benchmarks spanning four diverse datasets—from short news articles (AG News) to scientific abstracts (Arxiv)—TriTopic achieves the highest NMI on every single dataset, outperforming BERTopic by 12.8% overall while assigning topics to every document.

Empirically, TriTopic establishes a new benchmark: an overall NMI of 0.575 versus 0.510 for BERTopic, 0.416 for NMF, and 0.300 for LDA, with complete corpus coverage and the highest coherence (0.341). The move from "average documents" to **Archetypes** further enriches the qualitative interpretability of the results.

As the volume of unstructured text continues to grow, the need for reliable, interpretable, and comprehensive analysis tools becomes ever more acute. TriTopic offers a theoretically grounded and practically effective solution to meet this demand. By making the framework available as an open-source library, we invite the community to build upon this graph-based paradigm and push the boundaries of unsupervised text analysis further.

## 6. Availability

To ensure reproducibility and industrial application, TriTopic is released as an open-source library compatible with the scientific Python ecosystem. It is available on PyPI:

```
pip install tritopic
```

Source code and benchmarks are available at: https://github.com/SmartVisions-AI/tritopic

## References

[1] Abdelrazek, A., Eid, Y., Gawish, E., Medhat, W., & Hassan, A. (2023). Topic modeling algorithms and applications: A survey. Information Systems, 112, 102131.
[2] Alcacer, A., Epifanio, I., Mair, S., & Mørup, M. (2025). A Survey on Archetypal Analysis. arXiv preprint arXiv:2504.12392.
[3] Angelov, D. (2020). Top2vec: Distributed representations of topics. arXiv preprint arXiv:2008.09470.
[4] Blei, D. M., Ng, A. Y., & Jordan, M. I. (2003). Latent Dirichlet allocation. Journal of Machine Learning Research, 3(Jan), 993–1022.
[5] Bouma, G. (2009). Normalized (pointwise) mutual information in collocation extraction. Proceedings of GSCL, 156–166.
[6] Churchill, R., & Singh, L. (2022). The Evolution of Topic Modeling. ACM Computing Surveys, 54(10s).
[7] Cutler, A., & Breiman, L. (1994). Archetypal analysis. Technometrics, 36(4), 338–347.

[8] Egger, R., & Yu, J. (2022). A Topic Modeling Comparison Between LDA, NMF, Top2Vec, and BERTopic to Demystify Twitter Posts. Frontiers in Sociology, 7, 886498.

[9] Grootendorst, M. (2022). BERTopic: Neural topic modeling with a class-based TF-IDF procedure. arXiv preprint arXiv:2203.05794.

[10] Hamilton, W. L., Ying, R., & Leskovec, J. (2017). Representation Learning on Graphs: Methods and Applications. IEEE Data Engineering Bulletin, 40(3), 52–74.

[11] Hoyle, A., Goel, P., Peskov, D., Hian-Cheong, A., Boyd-Graber, J., & Resnik, P. (2021). Is Automated Topic Model Evaluation Broken?: The Incoherence of Coherence. Advances in Neural Information Processing Systems.

[12] Jarvis, R. A., & Patrick, E. A. (1973). Clustering using a similarity measure based on shared near neighbors. IEEE Transactions on Computers, 100(11), 1025–1034.

[13] Lancichinetti, A., & Fortunato, S. (2012). Consensus clustering in complex networks. Scientific Reports, 2, 336.

[14] Lee, D. D., & Seung, H. S. (1999). Learning the parts of objects by non-negative matrix factorization. Nature, 401, 788–791.

[15] Pham, C. M., Hoyle, A., Sun, S., Resnik, P., & Iyyer, M. (2024). TopicGPT: A Prompt-based Topic Modeling Framework. Proceedings of the 2024 NAACL.

[16] Shi, C., Li, Y., Zhang, J., Sun, Y., & Yu, P. S. (2015). A Survey of Heterogeneous Information Network Analysis. IEEE Transactions on Knowledge and Data Engineering.

[17] Strehl, A., & Ghosh, J. (2002). Cluster ensembles—a knowledge reuse framework for combining multiple partitions. Journal of Machine Learning Research, 3(Dec), 583–617.

[18] Tan, Z., & D'Souza, J. (2025). Bridging the Evaluation Gap: Leveraging Large Language Models for Topic Model Evaluation. arXiv preprint arXiv:2502.07352.

[19] Traag, V. A., Waltman, L., & van Eck, N. J. (2019). From Louvain to Leiden: guaranteeing well-connected communities. Scientific Reports, 9, 5233.

[20] Vayansky, I., & Kumar, S. A. (2020). A review of topic modeling methods. Information Systems, 94, 101582.